\pdfoutput=1

\documentclass[11pt]{article}

\usepackage{EACL2023}

\usepackage{times}
\usepackage{latexsym}

\usepackage{graphicx}
\usepackage{amsmath}
\usepackage{amsfonts}
\usepackage{booktabs}
\usepackage{multirow}
\usepackage{pifont}
\usepackage{xspace}
\usepackage[capitalise]{cleveref}
\usepackage[T1]{fontenc}

\usepackage[utf8]{inputenc}

\usepackage{microtype}

\usepackage{inconsolata}

\newcommand*{\eg}{e.g.\@\xspace}

\makeatletter
\newcommand*{\etc}{%
    \@ifnextchar{.}%
        {etc}%
        {etc.\@\xspace}%
}
\makeatother

\title{XQA-DST: Multi-Domain and Multi-Lingual Dialogue State Tracking}

\author{Han Zhou\textsuperscript{1,}\thanks{\quad Work done while at UCL.}
\qquad
Ignacio Iacobacci\textsuperscript{2}
\qquad
Pasquale Minervini\textsuperscript{3,4}
\\
\textsuperscript{1}University of Cambridge
\quad
\textsuperscript{2}Huawei Noah’s Ark Lab, London
\\ 
\textsuperscript{3}University of Edinburgh
\quad
\textsuperscript{4}University College London\\
\texttt{hz416@cam.ac.uk}\quad \texttt{ignacio.iacobacci@huawei.com}\quad
\texttt{p.minervini@ed.ac.uk}
}
\begin{document}
\maketitle
\begin{abstract}
Dialogue State Tracking (DST), a crucial component of task-oriented dialogue (ToD) systems, keeps track of all important information pertaining to dialogue history: filling slots with the most probable values throughout the conversation. Existing methods generally rely on a predefined set of values and struggle to generalise to previously unseen slots in new domains. To overcome these challenges, we propose a domain-agnostic extractive question answering (QA) approach with shared weights across domains. To disentangle the complex domain information in ToDs, we train our DST with a novel domain filtering strategy by excluding out-of-domain question samples. With an independent classifier that predicts the presence of multiple domains given the context, our model tackles DST by extracting spans in active domains. Empirical results demonstrate that our model can efficiently leverage domain-agnostic QA datasets by two-stage fine-tuning while being both domain-scalable and open-vocabulary in DST. It shows strong transferability by achieving zero-shot domain-adaptation results on MultiWOZ 2.1 with an average JGA of 36.7\%. It further achieves cross-lingual transfer with state-of-the-art zero-shot results, 66.2\% JGA from English to German and 75.7\% JGA from English to Italian on WOZ 2.0.
\end{abstract}

\section{Introduction}
\def\thefootnote{}\footnotetext{Code is available at \url{https://github.com/hanzhou032/xqa-dst}}Task-oriented dialogue systems are designed to provide natural conversation with users and assist them in achieving daily goals. With the growth of task-oriented dialogue systems, there is an increasing interest in supporting dialogues among many domains and languages to fit the users' demands. However, either modelling a multi-domain or multi-lingual dialogue system requires substantial data collected in real scenarios. This data acquisition procedure is extremely expensive, and it motivates us to resolve this challenge 
by leveraging dialogue data in rich-resource domains and languages via zero-shot transfer learning.

Dialogue State Tracking (DST) is crucial for accurately extracting user intents and goals over multiple turns within the dialogue. Based on the tracked dialogue states, the dialogue manager makes corresponding next actions with back-end results, where the accuracy of the DST becomes absolutely vital. With a fully predefined ontology, traditional approaches tackle the DST as a classification problem by enumerating every combination of slot-value pairs \citep{mrksic-etal-2017-neural,zhong-etal-2018-global}. Those approaches are strongly limited by their scalability, as some slots (\eg \emph{name}) have an unbounded set of slot values. Secondly, they are generally not flexible to unseen slot-value pairs, making them more difficult to adapt to zero-shot transfer learning. Moreover, a completely predefined ontology is hard to acquire and not scalable for ToD systems in real applications.

To overcome those challenges, 
we take inspiration from \citet{gao-etal-2020-machine} and investigate how DST can be tackled by extracting slot values from user utterances directly.
In this paper, we propose a domain-independent and transferable dialogue state tracker within an extractive question answering architecture. Our model is responsible for filling the slot value by recognising specially designed domain-slot prompts by span prediction, which extracts answers from the input utterance by predicting the token positions. In addition, 
we introduce a novel domain filtering strategy in training and an independent multi-domain classifier in evaluation such that we only ask slot questions that appear in predicted domains. For example, given \emph{hotel} as the current turn domain, all questions under the \emph{train} domain are filtered out as there is no overlap between them. This simple but effective filtering strategy significantly reduces the noise from unnecessary questions in both the training and evaluation phases. Furthermore, we study unexplored impacts of two-stage fine-tuning on DST transfer learning with mono-lingual and multi-lingual question answering datasets. 

We call the final model XQA-DST: \mbox{\textbf{X}LM-R based} \textbf{D}ialogue \textbf{S}tate \textbf{T}racker in \textbf{Q}uestion \textbf{A}nswering. Our main contributions are summarised below:

\begin{itemize}
    \item We introduce XQA-DST, a novel domain-independent and transferable dialogue state tracker inspired by extractive question answering models. The model is able to recognise slot values by reformulating the task as an answer to a designed domain-slot question prompt by span prediction, which extracts answers from the input utterance by predicting the token positions. 
    \item We enable XQA-DST on question answering by zero-shot domain adaptation scenarios, showing its transferability capabilities. The final model shows state-of-the-art domain adaptation performance with an average JGA of 36.7\% for five domains on MultiWOZ 2.1.
    \item We show that our model is capable of both domain adaptation and cross-lingual transfer learning. We demonstrate its cross-lingual transferability by achieving state-of-the-art zero-shot results, 66.2\% JGA from English to German and 75.7\% JGA from English to Italian on WOZ 2.0.
\end{itemize}
\section{Related Work}

Traditional dialogue state tracking approaches mostly rely on a predefined ontology. \citet{lee2019sumbt} implement a slot-utterance matching module that computes the similarity between the utterance and each slot-value pair. \citet{lai2020simple} use BERT \citep{devlin2018bert} as the context encoder and generate the relevance score for every pair. Recently, \citet{lin-etal-2021-knowledge} and \citet{feng-etal-2022-dynamic} include schema graph networks to utilise inter-slot relationships. However, their scalability is strongly limited by the availability of the predefined ontology and schema graphs. 

To improve %
efficiency, span prediction methods have been proposed to tackle DST so that the slot can be filled by directly addressing values in the context. \citet{heck-etal-2020-trippy} implement copy mechanisms, but they use independent span projection layers for each slot, which make their model incapable of inference in new domains. \citet{zhou2019multi} and \citet{gao-etal-2020-machine} formulate the DST as a question answering problem, and they prepare questions for asking the model to answer values for every slot. We differentiate from these approaches by disentangling the complex domain information from domain filtering and domain classification strategies.

Generative approaches \citep{wu-etal-2019-transferable, 2020MA} provide an alternative way to handle DST. \citet{li-etal-2021-zero} introduce a generative question answering approach, GPT2-m, that leverages an autoregressive language model. Similarly, \citet{, lin-etal-2021-zero, lin-etal-2021-leveraging} propose T5DST, and they study the impacts of slot descriptions and cross-task transfer on domain adaptation. \citet{lee-etal-2021-dialogue} reformulate DST as prompting states via schema descriptions from language models. Recent end-to-end dialogue models \citep{peng-etal-2021-soloist, su-etal-2022-multi} also show strong supervised performance on DST.

Cross-lingual transfer learning for DST aims to leverage the labelled data in rich-resource languages and transfer learned knowledge to low-resource languages. \citet{chen-etal-2018-xl} study this problem and propose the XL-NBT teacher-student framework. \citet{liu2020attention} introduce an Attention-informed Mixed-Language Training (AMLT) method to build code-switching training sentences. They study the effectiveness of multi-lingual pretrained language models, XLM \citep{conneau2019cross} and mBERT \citep{devlin2018bert}, with their AMLT approach. \citet{qin2020cosdaml} propose a data augmentation framework, which encourages cross-lingual alignment by fine-tuning mBERT on generated code-switching data. \citet{moghe-etal-2021-cross} introduce intermediate fine-tuning on parallel sentences to improve the cross-lingual DST. To the best of our knowledge, we are the first work that studies the effectiveness of 
a multi-lingual pretrained language model, XLM-R \citep{conneau-etal-2020-unsupervised}, on DST without implementing additional cross-lingual alignment strategies.
\section{Multi-Domain and Multi-Lingual DST}
\begin{figure*}
    \centering
    \includegraphics[width=\textwidth]{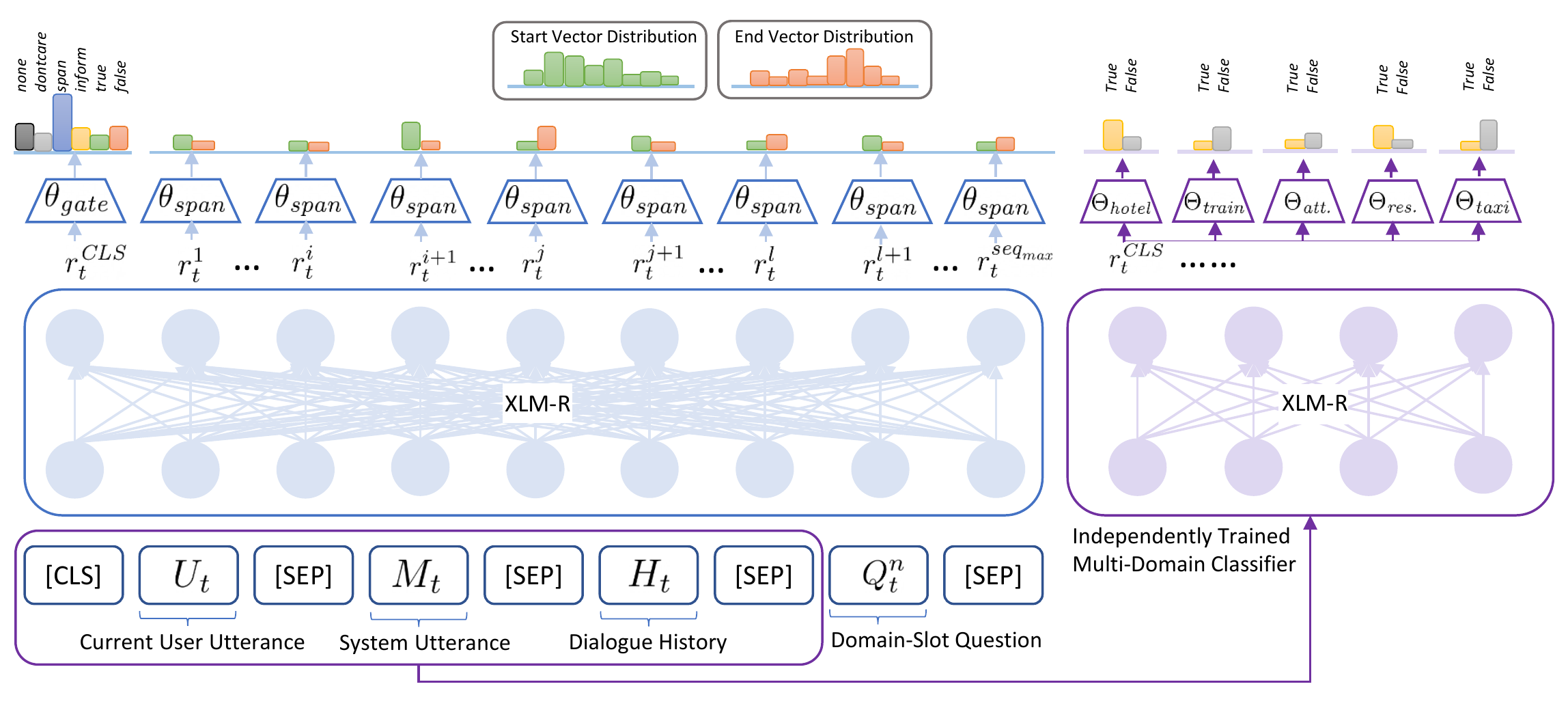}
    \caption{The model architecture of our XQA-DST for multi-domain and multi-lingual DST, where the right part is the independently trained multi-domain classifier that outputs active domains.}
    \label{fig:xlmrmsc}
\end{figure*}
To tackle the task of dialogue state tracking, our model reads the current user utterance $U_t$, preceding system utterance $M_t$, dialogue history $H_t$, and the domain-slot prompt $Q_t$ as inputs for each turn. Followed by that, our model is responsible for firstly determining the dialogue domains $D_t$ from the input sequence. Then, it predicts the presence of the answer span in predicted domains given the question. If an answer is present in utterances, the model will predict the value for that domain-slot question using span extraction. Otherwise, its value will be filled in accordance with other predicted types. Finally, our model tracks the dialogue states by a rule-based update mechanism along with the progress of the dialogue across turns.

\subsection{Context and Domain-Slot Questions}
In extractive question answering, the context is used to provide the background information, and the answer is usually contained in the context. When it comes to DST, it is equivalent to modelling the system message and the user response together as the context for the current turn. The complete context $C_t$ is then collected by concatenating the current user utterance $U_t$ and the preceding system utterance $M_t$ with dialogue history $H_t$ at turn $t$. We implement XLM-R as the context encoder for the purpose of cross-lingual transfer learning. 

Each context is paired with $N$ questions, which iterate through every slot that we are interested in. We append the domain-slot prompt at the end of the context as an analogue question for each domain-slot pair. Hence, the model can learn to correlate different questions to the same context and provide corresponding answers to fill the slot. For the same context with $n$th question $Q_t^n$ at turn $t$, the input sequence $S_t^n$ can be written as:
\begin{equation}
\begin{aligned}
        S_t^{n} = &[\text{CLS}] \oplus U_t \oplus [\text{SEP}] \oplus M_t \oplus [\text{SEP}] \\&\oplus H_t \oplus [\text{SEP}] \oplus Q_t^n \oplus [\text{SEP}], 
\end{aligned}
\label{Eq.history}
\end{equation}
where $\oplus$ is the string concatenation, and $H_t$ represents the dialogue history collected in a reversed order, and it is defined as follows:
\begin{equation}
\begin{aligned}
    H_t = & U_{t-1} \oplus M_{t-1} \oplus \dots 
    \\&\oplus U_{1} \oplus M_{1} \text{ for } t>1.
\end{aligned}
\end{equation}

To utilise the question as a distinct feature for each slot, we propose the analogue question in the format of a domain-slot prompt. Here, additional special tokens are introduced to assist the model in recognising the domain-slot pair as distinct parts. Moreover, they provide clear signals for the start and end positions for each domain-slot pair. The equation for constructing the domain-slot prompt $Q_t^n$ is defined below:
\begin{equation}
\begin{aligned}
    Q_t^n = &\langle \text{dom.} \rangle \oplus \textit{d}_t^n \oplus \langle /\text{dom.} \rangle 
    \\&\oplus \langle \text{slot} \rangle \oplus \textit{s}_t^n \oplus \langle \text{/slot} \rangle,
\end{aligned}
\label{Eq.domainquestion}
\end{equation}
where $d_t^n$ refers to the name of the domain and $s_t^n$ is the slot name for $n$-th question at turn $t$. 
\subsection{Shared Classification Gate}
Our model contains a shared classification gate $\theta_{gate}$ for every domain-slot question as shown in Fig. \ref{fig:xlmrmsc}. This shared gate provides shared knowledge among various domain-slot pairs, as it is neither domain-specific nor slot-specific. 

For each input sentence $S_t$, this shared gate classifies it to one of six classes as described in three main categories. Special cases, \textit{none/dontcare}, indicate that there is either no observable value from the input sequence $S_t$ or any value that can become the answer for that slot question. Copy mechanism, \textit{span}, indicates that the answer can be extracted from the current user utterance $U_t$ by the span prediction module. Similarly, \textit{Inform} is to copy from the system inform memory that tracks values mentioned in the preceding system utterance $M_t$. Boolean values \textit{true/false} are used to deal with binary categorical values for Boolean slots where the value cannot be extracted from the utterance. 

With these designed classes, it takes the output $r_t^{\mathrm{CLS}}$ from the encoder as its only input. It generates a probability distribution $p_{t}^{\mathrm{gate}} \in \mathbb{R}^{6}$ over six classes as in the following equation:
\begin{equation}
    p_t^{\mathrm{gate}} = \mathrm{softmax}(W_{\mathrm{gate}} \cdot r_t^{\mathrm{CLS}}+b_{\mathrm{gate}}),
\end{equation}
where $W_{\mathrm{gate}}$ represents the weights for our shared gate that is achieved by a linear classification layer, and $b_{\mathrm{gate}}$ is the corresponding bias term. The class is then determined by taking the maximal argument of $\mathrm{argmax}(p_t^{\mathrm{gate}})$.

\subsection{Shared Span Prediction Layer}
If the predicted class for the current input sequence $S_t$ is \textit{span}, the answer for that domain-slot question $Q_t$ will be filled by predicting the start and end positions of the value from the input sequence. We implement the shared span prediction layer for every domain-slot question for the purpose of domain-adaptable design. This is achieved by constructing a linear layer that takes the entire token representations from $r_t^1$ to $r_t^{\textit{seq}_{\textit{max}}}$ as inputs, and it generates two outputs with two parallel softmax layers for token positions, the start and end position distribution, $p_t^{\text{start}}$ and $p_t^{\text{end}}$.
\begin{subequations}
\begin{align}
    [p_t^{\text{start}}, p_t^{\text{end}}] &= \text{softmax}(W_{\text{span}} \cdot r_t^i + b_{\text{span}})\\
    \text{start}_t &= \text{argmax}(p_t^{\text{start}})\\
    \text{end}_t &= \text{argmax}(p_t^{\text{end}}).
\end{align}
\end{subequations}

The start and end positions of the predicted value are then determined by picking the largest probability from distributions $p_t^{\text{start}}$ and $p_t^{\text{end}}$. Followed by that, we sequentially collect the tokens from the predicted $\text{start}_t$ position to $\text{end}_t$ position, treating any reversed sequence prediction as an empty value. We then detokenize them to form the final predicted value for that domain-slot question.

\subsection{Turn-Domain Filtering}
For a task-oriented dialogue, the user may shift the domain of conversation across turns so that a dialogue can have multiple domains. We introduce a novel turn-domain filtering strategy that puts a strict constraint and only allows the model to pay attention to currently active domains. Turn-domain filtering indicates that only the slots within the current domains $D_t$ are used to prepare training features since slots are domain-specific. Hence, turn-domain filtering can reduce the potential noises introduced by unnecessary domains. Mathematically, this filtering strategy puts an additional constraint for slot domain $d_t^n$ in Eq. \ref{Eq.domainquestion}:
\begin{equation}
d_t^n\in D_t.
\end{equation}

\subsection{Independent Multi-Domain Classifier}
Turn-domain filtering allows the model to answer questions only within the interested domains. However, the domain information is no longer a given feature in the evaluation stage. Here, we propose a multi-domain sequence classifier as shown in Fig. \ref{fig:xlmrmsc}. The input sequence is the complete dialogue context $C_t$ without domain-slot questions. We then collect the entire sequence representation $r_t^{\text{CLS}}$ by the context encoders as $\text{XLM-R}(C_t)$. Followed by that, $r_t^{\text{CLS}}$ is fed into $|D|$ softmax layers, thereby allowing a binary prediction that decides whether each domain $d_t$ is present in the input context or not. Finally, we collect the domains that have been assigned to the `\textit{True}' class, which indicates the presence of that domain in the context. 
\begin{subequations}
\begin{gather}
        p_t^{d} = \text{softmax}(W_{\text{MSC}}^{d} \cdot r_t^{\text{CLS}} + b_{\text{MSC}}^d)\\
        d_t = \text{argmax}(p_t^d)\\
        D_t = \{d_1,\dots,d_{|D|}\}.
\end{gather}
\end{subequations}

\subsection{Inform Memory and Update Rules}
To further reduce the error of our span extractor, we have employed the same inform copy mechanism as \citet{heck-etal-2020-trippy}. This memory is a simple dictionary that records all values informed by the preceding system utterance $M_t$ into a system inform memory $I_t$=$\{I_t^1$, ..., $I_t^N\}$. Then, the value answer $A_t^n$ for $n$th question $Q_t^n$ asked at turn $t$ can be predicted by the following copy mechanism, given that $\textit{inform} = \text{argmax}(p_t^{\text{gate}})$:

\begin{equation}
    A_t^n = I_t^n \text{ for } Q_t^n.
\end{equation}

We implement a simple rule-based mechanism that is used to update dialogue states across turns as same as \citet{DBLP:conf/interspeech/ChaoL19}. In each turn, if the model assigned class for the current input sequence $S_t^n$ with $Q_t^n$ is not \textit{none}, the dialogue state will be updated by obtaining $A_t^n$ from our value prediction modules. On the other hand, if the classification gate predicts that there is no value for $S_t^n$, the dialogue state will be kept unchanged.

\subsection{Two-Stage Fine-Tuning}
Our model is designed to be capable of not only DST tasks but also general question answering tasks. Therefore, the transfer learning ability of our base model can be enhanced by firstly fine-tuning it on mono-lingual and multi-lingual question answering datasets as the first-stage fine-tuning. Then, we initialise its weights on DST shared gates and further fine-tune the model on DST datasets as the second-stage fine-tuning. This two-stage fine-tuning strategy maximally brings domain-agnostic knowledge into the field of DST. 
\section{Experimental Setup}
\subsection{Dataset}
The datasets that we carry out experiments on are WOZ 2.0 \citep{wen-etal-2017-network} and MultiWOZ 2.1 \citep{eric-etal-2020-multiwoz} for single-domain and multi-domain task-oriented dialogues, respectively. WOZ 2.0 is a restaurant reservation dataset, and it contains three slots: \textit{area}, \textit{food}, and \textit{price range}. It provides the conversation in three languages: English, German, and Italian. MultiWOZ 2.1 contains multi-domain conversations for more than 10000 dialogues over seven domains. The dialogue domain can change across turns, thereby making MultiWOZ 2.1 the most challenging dataset for task-oriented dialogue systems. We exclude \textit{hospital} and \textit{police} domains with very few dialogues, and the remaining dataset contains five domains (\textit{hotel}, \textit{train}, \textit{attraction}, \textit{restaurant}, and \textit{taxi}) with 30 domain-slot pairs in total. For domain adaptation experiments, we use an extractive QA dataset, SQuAD 2.0 \citep{rajpurkar-etal-2018-know}, to provide the intermediate fine-tuning. In cross-lingual experiments, we further use the multilingual QA dataset, XQuAD \citep{artetxe-etal-2020-cross}, to study the effectiveness of multi-lingual intermediate fine-tuning.

\begin{table*}
\centering
\setlength{\tabcolsep}{10pt}
\renewcommand{\arraystretch}{0.8}
\begin{tabular}{@{}lccccccc@{}}
\toprule
\textbf{Models} & \textbf{Type} &\textbf{Hotel} & \textbf{Train} & \textbf{Att.} & \textbf{Res.} & \textbf{Taxi} &\textbf{Avg.}\\ \midrule

MA-DST \citep{2020MA}& G                       & 16.3               & 22.8           & 22.5  &13.6&59.3&26.9\\
SUMBT \citep{lee2019sumbt}& C                  & 19.8      & 22.5           & 22.6           &16.5&59.5&28.2\\
TRADE \citep{wu-etal-2019-transferable}& G                        & 19.5               & 22.9           & 22.8  &16.4&59.2&28.2\\

GPT2-m \citep{li-etal-2021-zero} & G                  & \textbf{24.4}      & 29.1           & 31.3           &26.2&59.6&34.1\\

T5DST* \citep{lin-etal-2021-leveraging} & G                        & 21.2               & 35.4  & \textbf{33.1}           &21.7 & \textbf{64.6} & 35.2\\
TransferQA \citep{lin-etal-2021-zero} & G & 22.7 & 36.7 & 31.3 & 26.3 & 61.9 & 35.8\\
\midrule
XQA-DST \textit{w/o} two-stage& S                         & 22.9               & 37.0 & 24.0           &25.7&62.2&34.4\\
XQA-DST \textit{w.} SQuAD2  & S                        & 24.3               & \textbf{40.0}  & 27.9           & \textbf{28.2} & 63.2 & \textbf{36.7}\\
\bottomrule
\end{tabular}
\caption{The joint goal accuracy (\%) of zero-shot domain adaptation experiments on each domain with recent models on MultiWOZ 2.1. The abbreviations for model types are: G: Generative; C: Classification; S: Span prediction. *Results from MultiWOZ 2.0 are reported by \citet{lin-etal-2021-leveraging}.}
\label{Tabdaranks}
\end{table*}

\subsection{Implementation Details}
We employ the pretrained \textit{XLM-RoBERTa-base} model from the Huggingface library of Transformers \citep{wolf-etal-2020-transformers}, which consists of 12 hidden layers of 768 units. We also employ the \textit{BERT-base-uncased} model for ablation study and fine-tuned models on SQuAD 2.0 and XQuAD for adaptation experiments. For all implementations, we limit the maximal input sequence length to 180 tokens to save the cost while keeping a reasonable length for including dialogue history. We truncate from the earliest dialogue history when the input sequence length exceeds the limit. The training objective is to minimise the summations of individual loss functions for each module, where each loss is defined as the cross-entropy loss. The loss for each domain module in the multi-domain classifier is equally weighted, where the coefficient for each part of the joint loss of our main model is:
\begin{equation}
    \mathcal{L}_{\textit{total}}=0.8\cdot \mathcal{L}_{\textit{gate}}+0.2\cdot \mathcal{L}_{\textit{span}}.
\end{equation}

During the training process, we implement the Adam optimiser \citep{kingma2014adam} with an initial learning rate of $10^{-5}$. Then, we employ a linear scheduler with a warm-up proportion of $10\%$ so that the learning rate will decay linearly until reaching zero after the warm-up steps. We put a dropout layer with a rate of $30\%$ at the output of our context encoders. We use an early stopping strategy by monitoring the accuracy of the validation dataset until it stops increasing for at least 3 epochs. The batch size is fixed at $16$. The multi-domain classifier is trained independently with the same experimental setting, and it is only involved in the evaluation stage. We report the mean of supervised DST and zero-shot experimental results for three runs with different random seeds. 
\section{Experimental Results}

\subsection{Zero-Shot Domain Adaptation}
We rank our XQA-DST model with prior methods capable of zero-shot domain adaptation. The experiment is used to evaluate the transfer performance of models when tested with dialogues in a completely unseen domain. We train our model on the other four domains by excluding the target domains. We follow the experimental steps reported by \citet{2020MA}. Since there is a single domain defined in the target domain, the domain classifier is not utilised here because the dialogue domain is given information. Table \ref{Tabdaranks} shows a comparison of our XQA-DST model to baselines and recent approaches, where the JGA is defined as the ratio of dialogue turns that have been perfectly predicted over the number of turns for all dialogues. It is clear that our model has generated more accurate results than both MA-DST \citep{2020MA} and SUMBT \citep{lee2019sumbt} baselines by at least $6.2\%$ JGA on average in domain adaptation even without two-stage fine-tuning. SUMBT tracks the dialogue states by classifying every slot-value pair. Hence, it is a classification-based method, whereas our approach is mainly relying on the value filling by the span prediction module. It can be seen that our model has outperformed baselines by a significant (3-9\%) margin in the \textit{hotel}, \textit{restaurant}, and \textit{taxi} domains. This is because the classification-based method requires a predefined ontology for its enumeration of values, which inevitably makes it not robust to unseen values in new domains and results in relatively low performance for domain adaptation. 

There is another class of methods that utilises generative value filling to handle the DST, including TRADE, GPT2-m, and TransferQA. Given GPT2-m as an example, it is in the framework of generative question answering, which also coincides with the underlying idea of our XQA-DST model but has a decoder to generate candidate values. With the two-stage fine-tuning strategy on the SQuAD 2.0 dataset, our model shows improvements in all domains of 2.3\% on average. It shows the highest JGA in both \textit{train} and \textit{restaurant} domains (40.0\% and 28.2\%, respectively). It also outperforms the TransferQA approach that implements the cross-task transfer learning, which is similar to our two-stage fine-tuning that includes multi-task knowledge. Our results appear as the state-of-the-art results at 36.7\% JGA on average for zero-shot domain adaptation experiments.

Furthermore, our approach is designed to be applicable for both domain adaptation and cross-lingual transfer learning, whereas all generative methods listed above can only do mono-lingual learning. Therefore, our XQA-DST model has shown very competitive results in the zero-shot domain adaptation, and we can conclude that it is able to effectively generalise to task-oriented dialogues in new domains by understanding the linguistics behind our domain-slot questions.

\begin{figure}
    \centering
    \hspace{-0.9em}
    \includegraphics[width=\linewidth]{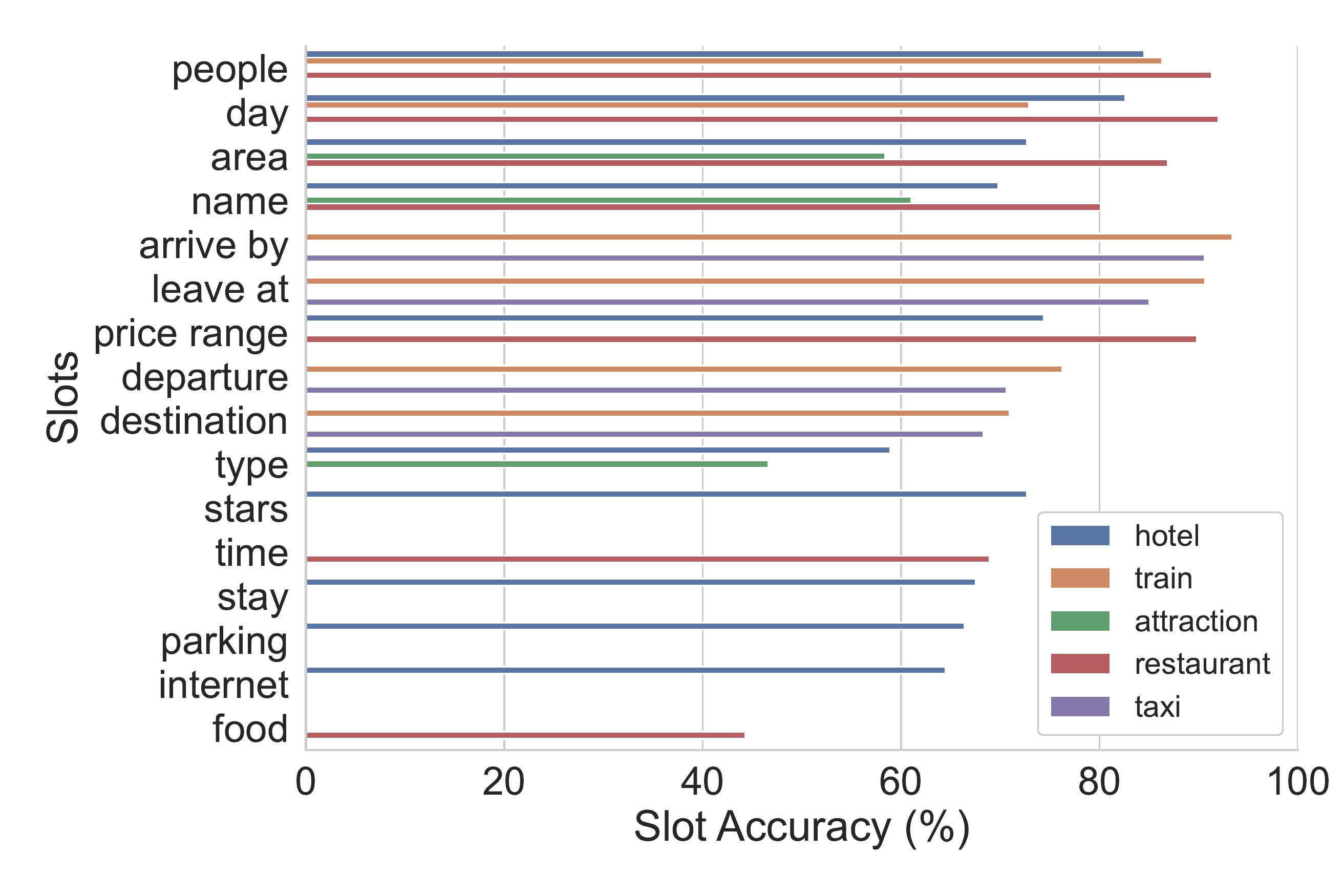}
    \caption{The categorical plot of slot accuracy (\%) for each slot over 5 domains for the zero-shot domain adaptation experiment by XQA-DST.}
    \label{fig:error}
\end{figure}

\subsection{Domain Adaptation Analysis}
\label{sec:erroranalysis}

We analyse the individual slot accuracy for every domain-slot pair in 5 domains to study the impact of shared slots over domains on the performance of domain adaptation. The results are obtained by computing the slot accuracy on each target domain by XQA-DST. The slot accuracy is defined as the ratio of dialogue turns where the value for that slot is correctly predicted. Fig. \ref{fig:error} shows the slot accuracy for 16 slots over 5 domains, where multiple domain bars for the same slot indicate that the slot is shared across these domains.

It is observable that slots that have been shared among multiple domains lead to a relatively higher domain adaptation performance. By contrast, it is also distinctive that slots that have not been shared among multiple domains have much lower accuracy. For instance, most slots in the \textit{hotel} domain are not shared with other domains, so the slot accuracy for  `\textit{parking}' and `\textit{internet}' slots (66.4\% and 64.5\%, respectively) are reasonably lower than others. The same rule applies to the `\textit{time}' and `\textit{food}' slots in the \textit{restaurant} domain. Therefore, the number of shared domains for the slot is the foremost factor in achieving a good domain adaptation result. Secondly, we notice that slots with digital values such as `\textit{people}' and `\textit{day}' have very high slot accuracy (91.3\% and 92.0\% in the \textit{restaurant} domain) even in the zero-shot setting. It validates the effectiveness of our model to domain adaptation for successfully extracting candidate values from the message. Last but not least, due to the wide surface form of location values, it is naturally hard to predict location slots, `\textit{departure}' and `\textit{destination}', that are not categorical with unseen values. Hence, even though they are shared in both \textit{train} and \textit{taxi} domains, they give 
relatively lower slot accuracy in the set of shared slots. Overall speaking, our XQA-DST model has generated reasonably well domain adaptation results on most domain-slot pairs and has shown a certain level of common knowledge across domains.

\begin{table}
\centering
\setlength{\tabcolsep}{10pt}
\renewcommand{\arraystretch}{1.1}
\small
\begin{tabular}{lcc}
\toprule

\multirow{2}{*}{\textbf{Models}} & \multicolumn{2}{c}{\textbf{Joint Goal Accuracy (\%)}} \\ 
  & \textbf{GE} & \textbf{IT} \\
 
\midrule

XLM-R-DST & 20.78 & 25.39 \\

\midrule

XL-NBT   & 30.80 & 41.20 \\
MUSE + AMLT   & 36.51 & 39.35 \\
XLM+CLCSA  & 48.70 & - \\
mBERT+CLCSA   & 63.20 & 61.30 \\
TLM+CLCSA & 65.80 & 66.90 \\
\midrule
Ours \textit{w/o} two-stage  & 64.88 & 68.63 \\
\hspace{2em} \textit{w.} XQuAD & \textbf{66.16} & 72.84 \\
\hspace{2em} \textit{w.} SQuAD2 & 66.12 & \textbf{75.66} \\

\bottomrule

\end{tabular}
\caption{The zero-shot cross-lingual DST results for target languages, German (GE) and Italian (IT), on WOZ 2.0. There are no results on Italian by XLM due to the absence of Italian in its pretraining \citep{liu2020attention}.}
\label{tab:zeroshotcl}
\end{table}

\subsection{Zero-Shot Cross-Lingual DST}
The zero-shot cross-lingual transfer learning is to train our XQA-DST on the source language, English. Then, it is sequentially evaluated on the test sets in German and Italian with labels that are kept in English. Since WOZ 2.0 is a single domain dataset with relatively short dialogues, the dialogue history is not included as inputs, and the domain classifier is deactivated. To provide a fair comparison to the ground truth, we implement Google Translator \citep{wu2016googles} to translate the values filled by span prediction in the target language back to the source language.

In Table \ref{tab:zeroshotcl}, our XQA-DST model with two-stage fine-tuning gives strong zero-shot results in both German and Italian languages (66.2\% and 75.7\% JGA, respectively). %
In comparison to recent approaches for cross-lingual DST, our XQA-DST model has generated results that significantly increase the margin by an absolute 8\% on Italian. It is worth noting that both XLM+CLCSA and mBERT+CLCSA \citep{qin2020cosdaml} are data augmentation-based approaches on multi-lingual models with the same model architecture as XL-NBT \citep{chen-etal-2018-xl}. TLM+CLCSA \citep{moghe-etal-2021-cross} also implements two-stage fine-tuning with data augmentation. Even without two-stage fine-tuning, our model in extractive QA still outperforms most of them and appears as the state-of-the-art results in the zero-shot cross-lingual transfer learning on WOZ 2.0. 

Besides the above approaches, we include XLM-R-DST as a baseline that we replace the context encoder of BERT-DST \citep{lai2020simple} with XLM-R. Then, we can study the effectiveness of different model architectures in cross-lingual transfer learning. We recall that XLM-R-DST fills the slot values by iterating through every candidate slot value with a relevance scorer. Table \ref{tab:zeroshotcl} shows a huge improvement in our approach by increasing the average JGA on target domains from 23.1\% to 66.8\% by more than 40\%. It indicates that our specially designed extractive QA framework has a strong generalisation ability across languages, whereas the XLM-R-DST appears as only recognising each value as distinct features without understanding the deep semantics behind them. Lastly, we notice that the cross-lingual result on Italian has a higher joint goal accuracy than German in our experiments. We suppose that this is because of the declension in German, which leads to more diverse word forms with the same semantics and introduces noises to the translation process.

\subsection{Supervised DST}
We perform experiments on the supervised DST configuration and compare our XQA-DST model with prior methods capable of monolingual zero-shot domain adaptation on MultiWOZ 2.1. Table \ref{tab:multiwozranks} comprises the JGA for each method, and we implement the same label mapping as TripPy \citep{heck-etal-2020-trippy} for a fair evaluation. In Table \ref{tab:multiwozranks}, our approach has outperformed most prior methods capable of zero-shot generalisation, including many generative approaches such as TRADE, T5DST, and GPT2-m. Though it is less competitive than the prompt-based SDP-DST model and end-to-end models in the supervised DST setting, its language transferability is still distinctive.

\begin{table}[t]
\setlength{\tabcolsep}{8pt}
\renewcommand{\arraystretch}{0.85}
\centering
\begin{tabular}{@{}lcccc@{}}
\toprule
\textbf{Models tested on MultiWOZ 2.1} & \textbf{JGA (\%)} \\ \midrule
TRADE \citep{wu-etal-2019-transferable} & 45.60 $\ $ \\
SUBMT \citep{lee2019sumbt} & 46.70 $\ $\\
STARC \citep{gao-etal-2020-machine}& 49.48 $\ $\\
MA-DST \citep{2020MA} & 51.88 $\ $\\
T5DST \citep{lin-etal-2021-leveraging}& 52.21 $\ $\\
GPT2-m \citep{li-etal-2021-zero}& 52.58 $\ $\\

SDP-DST \citep{lee-etal-2021-dialogue}& 56.66 $\ $\\
SOLOIST \citep{peng-etal-2021-soloist}& 56.85 $\ $\\
PPTOD \citep{su-etal-2022-multi}& \textbf{57.10} $\ $\\
\midrule
XQA-DST (our work)& 53.21 $\ $\\
\bottomrule
\end{tabular}
\caption{The performance of Supervised DST for our proposed XQA-DST model with prior methods capable of zero-shot inference on MultiWOZ 2.1.}
\label{tab:multiwozranks}
\end{table}

Based on the shared span prediction module, our model is able to extract values from the dialogue context directly, thereby being open-vocabulary and domain scalable. At the same time, it has successfully overcome the challenge of an unavailable ontology set. Moreover, it presents as the best-performed model in any framework with span prediction modules, where it has improved the margin of JGA by more than $3.5\%$ from the STARC approach. None of the other approaches has ever studied their DST with multi-lingual pretrained models. By utilising the pretrained XLM-R model as the context encoder, our approach is the only method with cross-lingual transferability. Given its distinct advantages of being domain-adaptable and language transferable, a promising result in multi-domain DST at $53.2\%$ is still competitive in the supervised setting.

To study the impact of essential designs in our model, we first analyse the performance of a mono-lingual model, BERT, and ablate it over different choices of domain classifiers. In Table \ref{tab:ablation}, the vanilla model with undersampling of negative samples has the lowest JGA at 38.2\%. This is because the shared span prediction layer lacks domain knowledge and frequently generates false positive predictions for out-of-domain questions. Introducing a joint domain classifier at the output of the main model in parallel with $\theta_{gate}$ improves the JGA by about 3\%, which convinces us about the effectiveness of domain classifiers. At the cost of the model size, the independent domain classifier significantly improves the JGA to 49.0\% by removing the interference from asking out-of-domain questions. It encourages the model to learn to distinguish in-domain questions rather than additionally learning the relationship between the context and domains within a goal. We notice that implementing XLM-R instead of BERT further improves the performance to 51.7\%. We suppose it is because of the well-trained RoBERTa model, and the multi-lingual pretraining does not greatly sacrifice the per-language performance. Lastly, due to the complexity of MultiWOZ dialogues, the history information is essential in accurately predicting current domains and extracting spans. Hence, appending the dialogue history has led our model to outperform most prior methods capable of zero-shot inference.

\begin{table}
\setlength{\tabcolsep}{4pt}
\renewcommand{\arraystretch}{0.85}
\centering
\begin{tabular}{@{}lcccc@{}}
\toprule
\textbf{Ablation} & \textbf{JGA (\%)} \\ \midrule
BERT-base   &  $\ $ \\
\hspace{2em} \textit{w.} undersampling & 38.23 $\ $\\
\hspace{2em} \textit{w.} joint domain classifier   & 41.10 $\ $\\
\hspace{2em} \textit{w.} independent domain classifier & 49.04 $\ $\\
\hspace{2em} \textit{+ } dialogue history  & 51.11 $\ $\\
XLM-RoBERTa-base &  $\ $\\
\hspace{2em} \textit{w.} independent domain classifier & 51.67 $\ $\\
\hspace{2em} \textit{+ } dialogue history  & 53.21 $\ $\\
\bottomrule
\end{tabular}
\caption{Ablation study of XQA-DST with different base models and training strategies on MultiWOZ 2.1.}
\label{tab:ablation}
\end{table}

\section{Conclusion}
We introduce a new multi-domain and multi-lingual dialogue state tracker, XQA-DST, within an extractive question answering framework. It gives distinct advantages for avoiding relying on any predefined ontology and being open-vocabulary to new slots with unseen values. We have shown a strong domain and cross-lingual transferable ability of our model by outperforming famous baselines. We have demonstrated its competitive performance in multi-domain DST with a novel turn-domain filtering strategy and a multi-domain classifier in parallel. With the design of an XLM-R based multi-domain classifier, our approach is feasible for tracking states in multi-domain and multi-lingual scenarios. Therefore, it holds a strong potential to overcome the challenging data scarcity problem for either domains or languages in the real application of task-oriented dialogue systems. 

\section*{Limitations}
In the supervised DST experiments, our multi-domain classifier is effective when the range of domains is given. However, we have fixed weights for each domain projection layer, which inevitably makes the classifier not domain scalable. Though the shared span prediction layer is still scalable to all domains, the performance of our model will degrade if it encounters a dialogue in multiple unseen domains. 

We recall that the independent multi-domain classifier provides a clearer training objective and significantly improves the JGA than the joint domain classifier. However, this is at the cost of model size and requires expensive computation resources. Therefore, we look forward to approaches that wisely incorporate the domain classifier. 

In the cross-lingual experiments, we test the transfer performance for German and Italian, which have been used as the pretraining languages for XLM-R. Hence, we expect a degradation of cross-lingual performance for our model on low-resource languages that are not pretrained by XLM-R. In addition, our experiments rely on a back-translation from the target language to the source language. Though we have implemented a predefined label dictionary that collects vocabulary with similar semantics, it cannot perfectly handle the noise from an external translation system. 

\paragraph{Acknowledgements}
Pasquale was partially funded by the European Union’s Horizon 2020 research and innovation programme under grant agreement no. 875160, ELIAI (The Edinburgh Laboratory for Integrated Artificial Intelligence) EPSRC (grant no. EP/W002876/1), an industry grant from Cisco, and a donation from Accenture LLP.

\bibliography{anthology,custom}

\begin{thebibliography}{32}
\expandafter\ifx\csname natexlab\endcsname\relax\def\natexlab#1{#1}\fi

\bibitem[{Artetxe et~al.(2020)Artetxe, Ruder, and
  Yogatama}]{artetxe-etal-2020-cross}
Mikel Artetxe, Sebastian Ruder, and Dani Yogatama. 2020.
\newblock \href {https://doi.org/10.18653/v1/2020.acl-main.421} {On the
  cross-lingual transferability of monolingual representations}.
\newblock In \emph{Proceedings of the 58th Annual Meeting of the Association
  for Computational Linguistics}, pages 4623--4637, Online. Association for
  Computational Linguistics.

\bibitem[{Chao and Lane(2019)}]{DBLP:conf/interspeech/ChaoL19}
Guan{-}Lin Chao and Ian~R. Lane. 2019.
\newblock \href {https://doi.org/10.21437/Interspeech.2019-1355} {{BERT-DST:}
  scalable end-to-end dialogue state tracking with bidirectional encoder
  representations from transformer}.
\newblock In \emph{{INTERSPEECH}}, pages 1468--1472. {ISCA}.

\bibitem[{Chen et~al.(2018)Chen, Chen, Su, Wang, Yu, Yan, and
  Wang}]{chen-etal-2018-xl}
Wenhu Chen, Jianshu Chen, Yu~Su, Xin Wang, Dong Yu, Xifeng Yan, and
  William~Yang Wang. 2018.
\newblock \href {https://doi.org/10.18653/v1/D18-1038} {{XL}-{NBT}: A
  cross-lingual neural belief tracking framework}.
\newblock In \emph{Proceedings of the 2018 Conference on Empirical Methods in
  Natural Language Processing}, pages 414--424, Brussels, Belgium. Association
  for Computational Linguistics.

\bibitem[{Conneau et~al.(2020)Conneau, Khandelwal, Goyal, Chaudhary, Wenzek,
  Guzm{\'a}n, Grave, Ott, Zettlemoyer, and
  Stoyanov}]{conneau-etal-2020-unsupervised}
Alexis Conneau, Kartikay Khandelwal, Naman Goyal, Vishrav Chaudhary, Guillaume
  Wenzek, Francisco Guzm{\'a}n, Edouard Grave, Myle Ott, Luke Zettlemoyer, and
  Veselin Stoyanov. 2020.
\newblock \href {https://doi.org/10.18653/v1/2020.acl-main.747} {Unsupervised
  cross-lingual representation learning at scale}.
\newblock In \emph{Proceedings of the 58th Annual Meeting of the Association
  for Computational Linguistics}, pages 8440--8451, Online. Association for
  Computational Linguistics.

\bibitem[{Conneau and Lample(2019)}]{conneau2019cross}
Alexis Conneau and Guillaume Lample. 2019.
\newblock \href
  {https://proceedings.neurips.cc/paper/2019/hash/c04c19c2c2474dbf5f7ac4372c5b9af1-Abstract.html}
  {Cross-lingual language model pretraining}.
\newblock In \emph{NeurIPS}, pages 7057--7067.

\bibitem[{Devlin et~al.(2019)Devlin, Chang, Lee, and
  Toutanova}]{devlin2018bert}
Jacob Devlin, Ming{-}Wei Chang, Kenton Lee, and Kristina Toutanova. 2019.
\newblock \href {https://doi.org/10.18653/v1/n19-1423} {{BERT:} pre-training of
  deep bidirectional transformers for language understanding}.
\newblock In \emph{{NAACL-HLT} {(1)}}, pages 4171--4186. Association for
  Computational Linguistics.

\bibitem[{Eric et~al.(2020)Eric, Goel, Paul, Sethi, Agarwal, Gao, Kumar, Goyal,
  Ku, and Hakkani-Tur}]{eric-etal-2020-multiwoz}
Mihail Eric, Rahul Goel, Shachi Paul, Abhishek Sethi, Sanchit Agarwal, Shuyang
  Gao, Adarsh Kumar, Anuj Goyal, Peter Ku, and Dilek Hakkani-Tur. 2020.
\newblock \href {https://aclanthology.org/2020.lrec-1.53} {{M}ulti{WOZ} 2.1: A
  consolidated multi-domain dialogue dataset with state corrections and state
  tracking baselines}.
\newblock In \emph{Proceedings of the 12th Language Resources and Evaluation
  Conference}, pages 422--428, Marseille, France. European Language Resources
  Association.

\bibitem[{Feng et~al.(2022)Feng, Lipani, Ye, Zhang, and
  Yilmaz}]{feng-etal-2022-dynamic}
Yue Feng, Aldo Lipani, Fanghua Ye, Qiang Zhang, and Emine Yilmaz. 2022.
\newblock \href {https://doi.org/10.18653/v1/2022.acl-long.10} {Dynamic schema
  graph fusion network for multi-domain dialogue state tracking}.
\newblock In \emph{Proceedings of the 60th Annual Meeting of the Association
  for Computational Linguistics (Volume 1: Long Papers)}, pages 115--126,
  Dublin, Ireland. Association for Computational Linguistics.

\bibitem[{Gao et~al.(2020)Gao, Agarwal, Jin, Chung, and
  Hakkani-Tur}]{gao-etal-2020-machine}
Shuyang Gao, Sanchit Agarwal, Di~Jin, Tagyoung Chung, and Dilek Hakkani-Tur.
  2020.
\newblock \href {https://doi.org/10.18653/v1/2020.nlp4convai-1.10} {From
  machine reading comprehension to dialogue state tracking: Bridging the gap}.
\newblock In \emph{Proceedings of the 2nd Workshop on Natural Language
  Processing for Conversational AI}, pages 79--89, Online. Association for
  Computational Linguistics.

\bibitem[{Heck et~al.(2020)Heck, van Niekerk, Lubis, Geishauser, Lin, Moresi,
  and Gasic}]{heck-etal-2020-trippy}
Michael Heck, Carel van Niekerk, Nurul Lubis, Christian Geishauser, Hsien-Chin
  Lin, Marco Moresi, and Milica Gasic. 2020.
\newblock \href {https://aclanthology.org/2020.sigdial-1.4} {{T}rip{P}y: A
  triple copy strategy for value independent neural dialog state tracking}.
\newblock In \emph{Proceedings of the 21th Annual Meeting of the Special
  Interest Group on Discourse and Dialogue}, pages 35--44, 1st virtual meeting.
  Association for Computational Linguistics.

\bibitem[{Kingma and Ba(2015)}]{kingma2014adam}
Diederik~P. Kingma and Jimmy Ba. 2015.
\newblock \href {http://arxiv.org/abs/1412.6980} {Adam: {A} method for
  stochastic optimization}.
\newblock In \emph{{ICLR} (Poster)}.

\bibitem[{Kumar et~al.(2020)Kumar, Ku, Goyal, Metallinou, and
  Hakkani{-}T{\"{u}}r}]{2020MA}
Adarsh Kumar, Peter Ku, Anuj~Kumar Goyal, Angeliki Metallinou, and Dilek
  Hakkani{-}T{\"{u}}r. 2020.
\newblock \href {https://ojs.aaai.org/index.php/AAAI/article/view/6322}
  {{MA-DST:} multi-attention-based scalable dialog state tracking}.
\newblock In \emph{{AAAI}}, pages 8107--8114. {AAAI} Press.

\bibitem[{Lai et~al.(2020)Lai, Tran, Bui, and Kihara}]{lai2020simple}
Tuan~Manh Lai, Quan~Hung Tran, Trung Bui, and Daisuke Kihara. 2020.
\newblock \href {https://doi.org/10.1109/ICASSP40776.2020.9053975} {A simple
  but effective bert model for dialog state tracking on resource-limited
  systems}.
\newblock In \emph{{ICASSP}}, pages 8034--8038. {IEEE}.

\bibitem[{Lee et~al.(2021)Lee, Cheng, and Ostendorf}]{lee-etal-2021-dialogue}
Chia-Hsuan Lee, Hao Cheng, and Mari Ostendorf. 2021.
\newblock \href {https://doi.org/10.18653/v1/2021.emnlp-main.404} {Dialogue
  state tracking with a language model using schema-driven prompting}.
\newblock In \emph{Proceedings of the 2021 Conference on Empirical Methods in
  Natural Language Processing}, pages 4937--4949, Online and Punta Cana,
  Dominican Republic. Association for Computational Linguistics.

\bibitem[{Lee et~al.(2019)Lee, Lee, and Kim}]{lee2019sumbt}
Hwaran Lee, Jinsik Lee, and Tae{-}Yoon Kim. 2019.
\newblock \href {https://doi.org/10.18653/v1/p19-1546} {{SUMBT:} slot-utterance
  matching for universal and scalable belief tracking}.
\newblock In \emph{{ACL} {(1)}}, pages 5478--5483. Association for
  Computational Linguistics.

\bibitem[{Li et~al.(2021)Li, Cao, Sridhar, Zhu, Li, Hamza, and
  McAuley}]{li-etal-2021-zero}
Shuyang Li, Jin Cao, Mukund Sridhar, Henghui Zhu, Shang-Wen Li, Wael Hamza, and
  Julian McAuley. 2021.
\newblock \href {https://doi.org/10.18653/v1/2021.eacl-main.91} {Zero-shot
  generalization in dialog state tracking through generative question
  answering}.
\newblock In \emph{Proceedings of the 16th Conference of the European Chapter
  of the Association for Computational Linguistics: Main Volume}, pages
  1063--1074, Online. Association for Computational Linguistics.

\bibitem[{Lin et~al.(2021{\natexlab{a}})Lin, Tseng, and
  Byrne}]{lin-etal-2021-knowledge}
Weizhe Lin, Bo-Hsiang Tseng, and Bill Byrne. 2021{\natexlab{a}}.
\newblock \href {https://doi.org/10.18653/v1/2021.emnlp-main.620}
  {Knowledge-aware graph-enhanced {GPT}-2 for dialogue state tracking}.
\newblock In \emph{Proceedings of the 2021 Conference on Empirical Methods in
  Natural Language Processing}, pages 7871--7881, Online and Punta Cana,
  Dominican Republic. Association for Computational Linguistics.

\bibitem[{Lin et~al.(2021{\natexlab{b}})Lin, Liu, Madotto, Moon, Zhou, Crook,
  Wang, Yu, Cho, Subba, and Fung}]{lin-etal-2021-zero}
Zhaojiang Lin, Bing Liu, Andrea Madotto, Seungwhan Moon, Zhenpeng Zhou, Paul
  Crook, Zhiguang Wang, Zhou Yu, Eunjoon Cho, Rajen Subba, and Pascale Fung.
  2021{\natexlab{b}}.
\newblock \href {https://doi.org/10.18653/v1/2021.emnlp-main.622} {Zero-shot
  dialogue state tracking via cross-task transfer}.
\newblock In \emph{Proceedings of the 2021 Conference on Empirical Methods in
  Natural Language Processing}, pages 7890--7900, Online and Punta Cana,
  Dominican Republic. Association for Computational Linguistics.

\bibitem[{Lin et~al.(2021{\natexlab{c}})Lin, Liu, Moon, Crook, Zhou, Wang, Yu,
  Madotto, Cho, and Subba}]{lin-etal-2021-leveraging}
Zhaojiang Lin, Bing Liu, Seungwhan Moon, Paul Crook, Zhenpeng Zhou, Zhiguang
  Wang, Zhou Yu, Andrea Madotto, Eunjoon Cho, and Rajen Subba.
  2021{\natexlab{c}}.
\newblock \href {https://doi.org/10.18653/v1/2021.naacl-main.448} {Leveraging
  slot descriptions for zero-shot cross-domain dialogue {S}tate{T}racking}.
\newblock In \emph{Proceedings of the 2021 Conference of the North American
  Chapter of the Association for Computational Linguistics: Human Language
  Technologies}, pages 5640--5648, Online. Association for Computational
  Linguistics.

\bibitem[{Liu et~al.(2020)Liu, Winata, Lin, Xu, and Fung}]{liu2020attention}
Zihan Liu, Genta~Indra Winata, Zhaojiang Lin, Peng Xu, and Pascale Fung. 2020.
\newblock \href {https://ojs.aaai.org/index.php/AAAI/article/view/6362}
  {Attention-informed mixed-language training for zero-shot cross-lingual
  task-oriented dialogue systems}.
\newblock In \emph{{AAAI}}, pages 8433--8440. {AAAI} Press.

\bibitem[{Moghe et~al.(2021)Moghe, Steedman, and Birch}]{moghe-etal-2021-cross}
Nikita Moghe, Mark Steedman, and Alexandra Birch. 2021.
\newblock \href {https://doi.org/10.18653/v1/2021.emnlp-main.87} {Cross-lingual
  intermediate fine-tuning improves dialogue state tracking}.
\newblock In \emph{Proceedings of the 2021 Conference on Empirical Methods in
  Natural Language Processing}, pages 1137--1150, Online and Punta Cana,
  Dominican Republic. Association for Computational Linguistics.

\bibitem[{Mrk{\v{s}}i{\'c} et~al.(2017)Mrk{\v{s}}i{\'c}, {\'O}~S{\'e}aghdha,
  Wen, Thomson, and Young}]{mrksic-etal-2017-neural}
Nikola Mrk{\v{s}}i{\'c}, Diarmuid {\'O}~S{\'e}aghdha, Tsung-Hsien Wen, Blaise
  Thomson, and Steve Young. 2017.
\newblock \href {https://doi.org/10.18653/v1/P17-1163} {Neural belief tracker:
  Data-driven dialogue state tracking}.
\newblock In \emph{Proceedings of the 55th Annual Meeting of the Association
  for Computational Linguistics (Volume 1: Long Papers)}, pages 1777--1788,
  Vancouver, Canada. Association for Computational Linguistics.

\bibitem[{Peng et~al.(2021)Peng, Li, Li, Shayandeh, Liden, and
  Gao}]{peng-etal-2021-soloist}
Baolin Peng, Chunyuan Li, Jinchao Li, Shahin Shayandeh, Lars Liden, and
  Jianfeng Gao. 2021.
\newblock \href {https://doi.org/10.1162/tacl_a_00399} {Soloist: Building task
  bots at scale with transfer learning and machine teaching}.
\newblock \emph{Transactions of the Association for Computational Linguistics},
  9:807--824.

\bibitem[{Qin et~al.(2020)Qin, Ni, Zhang, and Che}]{qin2020cosdaml}
Libo Qin, Minheng Ni, Yue Zhang, and Wanxiang Che. 2020.
\newblock \href {https://doi.org/10.24963/ijcai.2020/533} {Cosda-ml:
  Multi-lingual code-switching data augmentation for zero-shot cross-lingual
  {NLP}}.
\newblock In \emph{{IJCAI}}, pages 3853--3860. ijcai.org.

\bibitem[{Rajpurkar et~al.(2018)Rajpurkar, Jia, and
  Liang}]{rajpurkar-etal-2018-know}
Pranav Rajpurkar, Robin Jia, and Percy Liang. 2018.
\newblock \href {https://doi.org/10.18653/v1/P18-2124} {Know what you don{'}t
  know: Unanswerable questions for {SQ}u{AD}}.
\newblock In \emph{Proceedings of the 56th Annual Meeting of the Association
  for Computational Linguistics (Volume 2: Short Papers)}, pages 784--789,
  Melbourne, Australia. Association for Computational Linguistics.

\bibitem[{Su et~al.(2022)Su, Shu, Mansimov, Gupta, Cai, Lai, and
  Zhang}]{su-etal-2022-multi}
Yixuan Su, Lei Shu, Elman Mansimov, Arshit Gupta, Deng Cai, Yi-An Lai, and
  Yi~Zhang. 2022.
\newblock \href {https://doi.org/10.18653/v1/2022.acl-long.319} {Multi-task
  pre-training for plug-and-play task-oriented dialogue system}.
\newblock In \emph{Proceedings of the 60th Annual Meeting of the Association
  for Computational Linguistics (Volume 1: Long Papers)}, pages 4661--4676,
  Dublin, Ireland. Association for Computational Linguistics.

\bibitem[{Wen et~al.(2017)Wen, Vandyke, Mrk{\v{s}}i{\'c}, Ga{\v{s}}i{\'c},
  Rojas-Barahona, Su, Ultes, and Young}]{wen-etal-2017-network}
Tsung-Hsien Wen, David Vandyke, Nikola Mrk{\v{s}}i{\'c}, Milica
  Ga{\v{s}}i{\'c}, Lina~M. Rojas-Barahona, Pei-Hao Su, Stefan Ultes, and Steve
  Young. 2017.
\newblock \href {https://aclanthology.org/E17-1042} {A network-based end-to-end
  trainable task-oriented dialogue system}.
\newblock In \emph{Proceedings of the 15th Conference of the {E}uropean Chapter
  of the Association for Computational Linguistics: Volume 1, Long Papers},
  pages 438--449, Valencia, Spain. Association for Computational Linguistics.

\bibitem[{Wolf et~al.(2020)Wolf, Debut, Sanh, Chaumond, Delangue, Moi, Cistac,
  Rault, Louf, Funtowicz, Davison, Shleifer, von Platen, Ma, Jernite, Plu, Xu,
  Le~Scao, Gugger, Drame, Lhoest, and Rush}]{wolf-etal-2020-transformers}
Thomas Wolf, Lysandre Debut, Victor Sanh, Julien Chaumond, Clement Delangue,
  Anthony Moi, Pierric Cistac, Tim Rault, Remi Louf, Morgan Funtowicz, Joe
  Davison, Sam Shleifer, Patrick von Platen, Clara Ma, Yacine Jernite, Julien
  Plu, Canwen Xu, Teven Le~Scao, Sylvain Gugger, Mariama Drame, Quentin Lhoest,
  and Alexander Rush. 2020.
\newblock \href {https://doi.org/10.18653/v1/2020.emnlp-demos.6} {Transformers:
  State-of-the-art natural language processing}.
\newblock In \emph{Proceedings of the 2020 Conference on Empirical Methods in
  Natural Language Processing: System Demonstrations}, pages 38--45, Online.
  Association for Computational Linguistics.

\bibitem[{Wu et~al.(2019)Wu, Madotto, Hosseini-Asl, Xiong, Socher, and
  Fung}]{wu-etal-2019-transferable}
Chien-Sheng Wu, Andrea Madotto, Ehsan Hosseini-Asl, Caiming Xiong, Richard
  Socher, and Pascale Fung. 2019.
\newblock \href {https://doi.org/10.18653/v1/P19-1078} {Transferable
  multi-domain state generator for task-oriented dialogue systems}.
\newblock In \emph{Proceedings of the 57th Annual Meeting of the Association
  for Computational Linguistics}, pages 808--819, Florence, Italy. Association
  for Computational Linguistics.

\bibitem[{Wu et~al.(2016)Wu, Schuster, Chen, Le, Norouzi, Macherey, Krikun,
  Cao, Gao, Macherey, Klingner, Shah, Johnson, Liu, Kaiser, Gouws, Kato, Kudo,
  Kazawa, Stevens, Kurian, Patil, Wang, Young, Smith, Riesa, Rudnick, Vinyals,
  Corrado, Hughes, and Dean}]{wu2016googles}
Yonghui Wu, Mike Schuster, Zhifeng Chen, Quoc~V. Le, Mohammad Norouzi, Wolfgang
  Macherey, Maxim Krikun, Yuan Cao, Qin Gao, Klaus Macherey, Jeff Klingner,
  Apurva Shah, Melvin Johnson, Xiaobing Liu, Lukasz Kaiser, Stephan Gouws,
  Yoshikiyo Kato, Taku Kudo, Hideto Kazawa, Keith Stevens, George Kurian,
  Nishant Patil, Wei Wang, Cliff Young, Jason Smith, Jason Riesa, Alex Rudnick,
  Oriol Vinyals, Greg Corrado, Macduff Hughes, and Jeffrey Dean. 2016.
\newblock \href {http://arxiv.org/abs/1609.08144} {Google's neural machine
  translation system: Bridging the gap between human and machine translation}.
\newblock \emph{CoRR}, abs/1609.08144.

\bibitem[{Zhong et~al.(2018)Zhong, Xiong, and Socher}]{zhong-etal-2018-global}
Victor Zhong, Caiming Xiong, and Richard Socher. 2018.
\newblock \href {https://doi.org/10.18653/v1/P18-1135} {Global-locally
  self-attentive encoder for dialogue state tracking}.
\newblock In \emph{Proceedings of the 56th Annual Meeting of the Association
  for Computational Linguistics (Volume 1: Long Papers)}, pages 1458--1467,
  Melbourne, Australia. Association for Computational Linguistics.

\bibitem[{Zhou and Small(2019)}]{zhou2019multi}
Li~Zhou and Kevin Small. 2019.
\newblock \href {https://arxiv.org/abs/1911.06192} {Multi-domain dialogue state
  tracking as dynamic knowledge graph enhanced question answering}.
\newblock In \emph{The third Conversational AI workshop @ NeurIPS 2019,
  Vancouver, BC, Canada, 13 December 2019}.

\end{thebibliography}
\bibliographystyle{acl_natbib}

\appendix

\section{Reproducibility Details}
\label{sec:appendix}
\paragraph{Training details}
We use both \textit{XLM-RoBERTa-base} (125M) and \textit{BERT-base-uncased} (110M) with pretrained weights from the Huggingface library of Transformers. We run all experiments on a single RTX 3080Ti with 12 GB memory. We fix the batch size at 16 for all models during training and use the batch size at 1 for evaluations. During training, it takes about 40 minutes to run an epoch on MultiWOZ 2.1, and its inference time for all evaluation examples is about 7 minutes. For the WOZ 2.0 dataset, it takes roughly 20 minutes to train the model. In the cross-lingual setting, the inference time is about 10 minutes due to the back-translation procedure.  
\paragraph{Hyperparameters}
For two-stage fine-tuning experiments, we implement QA fine-tuned models from the Huggingface library of Transformers without tuning their hyperparameters. For the XQuAD experiment, it implements the batch size at 40 and a learning rate of $3\times10^{-5}$ for the first-stage fine-tuning. For the SQuAD 2.0 experiment, we use the fine-tuned weights and hyperparameters from \textit{deepset/xlm-roberta-base-squad2}.
\paragraph{Dataset details}
For the supervised DST experiments, we split the datasets into train/dev/test sets as same as \citet{heck-etal-2020-trippy}. In domain adaptation experiments, the MultiWOZ 2.1 datasets are divided into 5 domains in accordance with \citet{lin-etal-2021-leveraging}, where the \textit{hospital} and \textit{police} domains are excluded. Lastly, the multi-lingual WOZ 2.0 datasets have the same split as \citet{moghe-etal-2021-cross}.
\end{document}